\documentclass[10pt]{llncs}
\usepackage{graphicx}
\usepackage{amsmath,amssymb} 
\usepackage{color}
\usepackage[width=122mm,left=12mm,paperwidth=146mm,height=193mm,top=12mm,paperheight=217mm]{geometry}
\begin{document}
\pagestyle{headings}
\mainmatter

\title{Patch DCT v.s. leNet}

\author{D.A.Sinclair}
\institute{david@imense.com, Imense Ltd. 
\\28th Oct 2022}

\maketitle

\begin{abstract}
  This paper compares the performance of a NN taking the output of a DCT (Discrete Cosine Transform)
  of an image patch with leNet for classifying MNIST hand written digits. The basis functions underlying the DCT bear
  a passing resemblance to some of the learned basis function of the Visual Transformer but are an order of
  magnitude faster to apply. 
  
\end{abstract}

\section{Introduction}
\label{sec:intro}
It is probably worth stating what this paper is not about to avoid any confusion. This paper is
not a review paper of OCR methods or of feature descriptors for image patches round SIFT style features \cite{DBLP:journals/pami/MikolajczykS05,gauglitz2011evaluation}.
It is not suggesting that DCT coefficients of image patches be used for OCR. What it does is to provide
a simple performance comparison of a fast basis function method against the standard CNN over a recognised data-set.
It is asking the question, is there a fundamental reason for CNNs or Visual Transformer methods
to be superior to each other? 

Convolution nets have represented the state of the art in visual classification \cite{726791} for a number
of years however larger networks with greater capacity have started to challenge this position.

The Visual Transformer \cite{DBLP:journals/corr/abs-2010-11929} in particular offers a different approach
using a higher dimensional model consisting of
embedded feature vectors derived from learned basis functions for sub-patches covering an extended image patch.
The most interesting aspect of the ViT \cite{DBLP:journals/corr/abs-2010-11929} for the purposes of this paper
is the form of the basis vectors used to map an image patch into the high dimensional space used for classification.
To the casual observer some of these basis vectors bear at least a passing resemblance to some of the basis
functions of the DCT \cite{DCT_url}.

The principle disadvantages of the ViT are the large training sets (300M images to train 90M weights) required to
create the basis vectors for mapping image content to the high dimensional feature space and the consequent
time required to evaluate the network. This paper implicitly asks the question can the learned basis vectors
of the ViT be replaced by a DCT without great loss in performance.

To provide a partial answer to this question this paper compares the performance of leNet (as defined on PyTorch.org)
and a simple NN model applied to the coefficients returned by the DCT when it is applied to an image patch.
The training and evaluation set is the familiar MNIST hand written digit data set \cite{lecun-mnist-2010}.
DCT coefficients have been used as direct features in OCR \cite{DCT_Gujarati} with the focus being on reducing
feature dimensionality for classifiers rather than training a MPL, this is likely because the authors were working
on obscure language recognition and they may not have had a large volume of labelled training data.

The paper is laid out as follows: section \ref{sec:DCT_model} details the models used and how they were trained.
Full code for building and training the models in PyTorch is given in appendix\ref{sec:pytorch_def}. 
Section \ref{subsec:patch_DCT} lays out the performance benefits of moving from multiple linear basis function to the DCT. 
  
\section{The patch DCT model}
\label{sec:DCT_model}

All of the models used in this paper are defined and trained in PyTorch. The training set used in this
paper was the MNIST hand written digit data set \cite{lecun-mnist-2010}
comprising 60k training samples and 10k test samples.
The samples were resacled to be 32x32 pixel and brightness range was scaled to $[0, 1.0]$.
Figure \ref{fig:MNIST} shows the first 12 characters of the MNIST data-set.

\subsection{leNet model}
\label{subsec:score}

The model for the leNet style CNN was taken directly form PyTorch.org. The code for the leNet network definition is:

\begin{verbatim}
class Net(nn.Module):

    def __init__(self):
        super(Net, self).__init__()
        self.conv1 = nn.Conv2d(1, 6, 5)
        self.conv2 = nn.Conv2d(6, 16, 5)
        self.fc1 = nn.Linear(16 * 5 * 5, 120)  
        self.fc2 = nn.Linear(120, 84)
        self.fc3 = nn.Linear(84, 10)

    def forward(self, x):
        x = F.max_pool2d(F.relu(self.conv1(x)), (2, 2))
        x = F.max_pool2d(F.relu(self.conv2(x)), 2)
        x = torch.flatten(x, 1) 
        x = F.relu(self.fc1(x))
        x = F.relu(self.fc2(x))
        x = self.fc3(x)
        return x

criterion = nn.MSELoss()
optimizer = optim.SGD(net.parameters(), lr=0.01)

\end{verbatim}

The network was trained over 500,000 training cycles with a batch size of 16 random samples to
give an overall accuracy on the test set of $ 98.94\% $ correct.

For interest figure \ref{fig:conv_kernels} shows the first layer convolution kernels learned
in the training process and their application to the number 3.
It is likely that greater generalisation accuracy could have been achieved through standard
data multiplication methods on the training set.

\subsection{The patch DCT model}
\label{subsec:patch_DCT}

The discrete cosine transform is defined to be:
\begin{equation}
   D_{pq} = \alpha_{p}\alpha_{q}\sum_{m=0}^{M-1}\sum_{n=0}^{N-1}I_{mn}\cos(\frac{\pi(2m+1)p}{2M})\cos(\frac{\pi(2n+1)p}{2N})
    \label{eq:DCT}
\end{equation}
\begin{equation}
    \alpha_{p} = 
    \left\{
        \begin{array}{lr}
            1 / \sqrt{M},  p = 0 \\
            \sqrt{2/M},  1 \leq p \leq M-1 
        \end{array}
    \right.
    \label{eq:aM}
\end{equation}
\begin{equation}
    \alpha_{q} = 
    \left\{
        \begin{array}{lr}
            1 / \sqrt{N},  q = 0 \\
            \sqrt{2/N},  1 \leq q \leq N-1 
        \end{array}
    \right.
    \label{eq:aN}
\end{equation}
where $M$ and $N$ represent image patch dimensions in pixels and $D_{pq}$
represents the coefficients of the DCT.
The separable nature of the DCT means that for square image patches
the transform can be executed as a matrix multiplication:
\begin{equation}
  D = T * I * T'
    \label{eq:DCT_m}
\end{equation}
where
\begin{equation}
    T_{pq} = 
    \left\{
        \begin{array}{lr}
            \frac{1}{ \sqrt{M}},  p = 0, 0 \leq q \leq M-1 \\
            \sqrt{\frac{2}{M}}cos\frac{m(2q+1)p}{2M},  1 \leq p \leq M-1,   0 \leq q \leq M-1
        \end{array}
    \right.
    \label{eq:aM_m}
\end{equation}

Using a DCT to estimate linear coefficients of a set of basis functions offers a significant computational
saving. For a 32x32 patch if a general spanning set of basis vectors is used then $ (32*32)^{2} = 10^{6} $ multiplications
are required. Using the matrix formulation of the DCT requires $ 2*32^{3} = 6.55*10^{4} $ multiplications.
For a NxN patch the DCT will take a factor of $ \frac{2}{N} $ fewer operations.

\begin{figure}[htbp]
    \centering
    \includegraphics[width=0.3\linewidth]{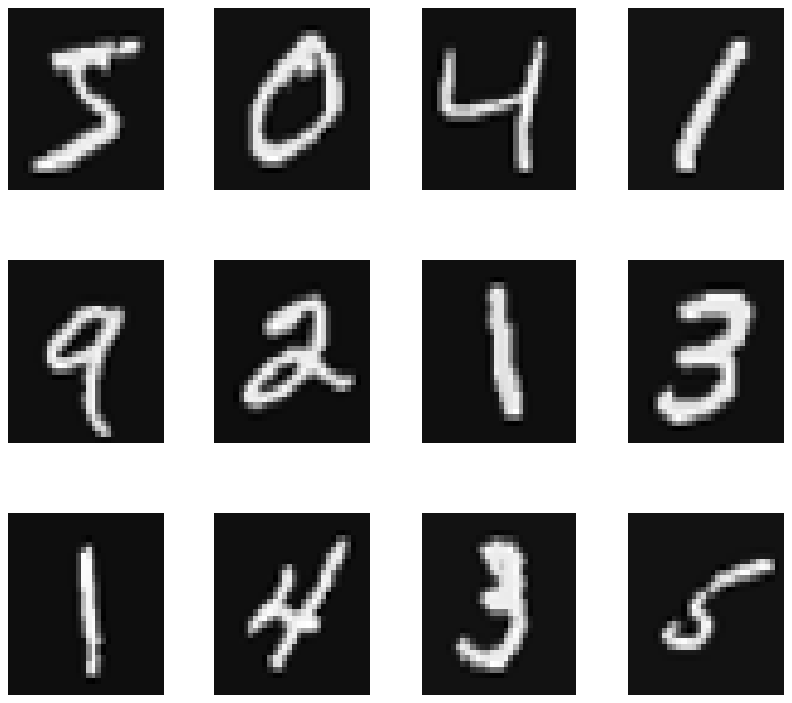} . 
    \includegraphics[width=0.3\linewidth]{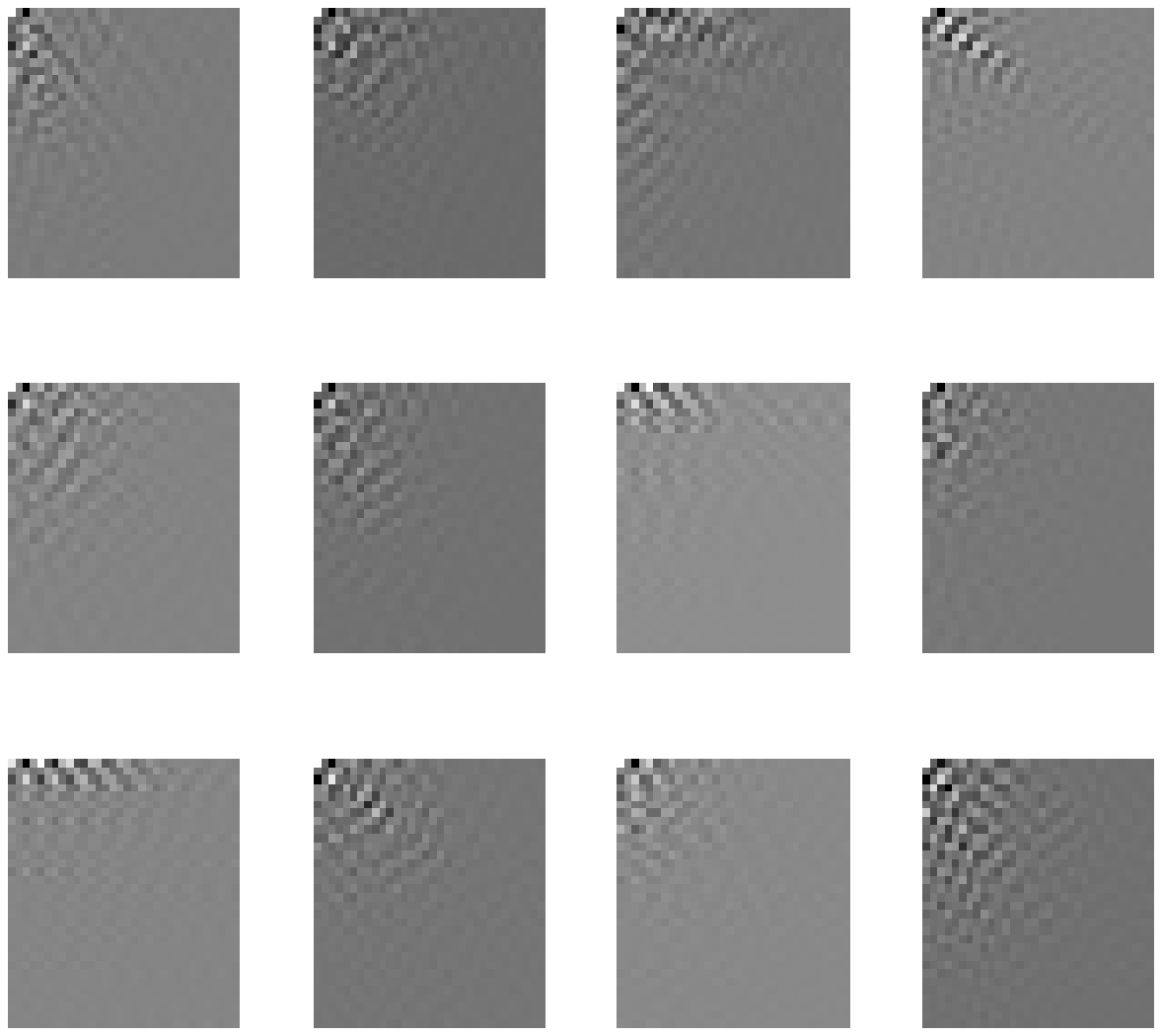}
    \caption{First 12 hand written digits form the MNIST data-set \cite{lecun-mnist-2010} and the corresponding DCT coefficients.}
    \label{fig:fig_1}
\end{figure}

\begin{figure}[htbp]
    \centering
    \includegraphics[width=0.7\linewidth]{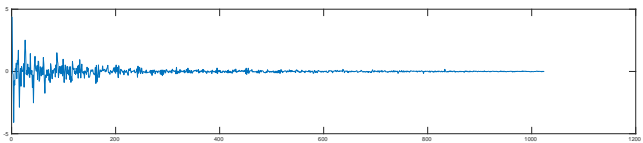} 
    \caption{Diagonally rastered DCT values for the first character in figure \ref{fig:fig_1}.}
    \label{fig:mag_plot}
\end{figure}

\begin{figure}[htbp]
    \centering
    \includegraphics[width=0.7\linewidth]{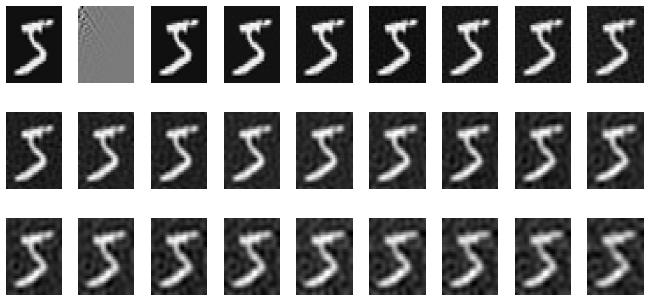} 
    \caption{Reconstructions of the first character in figure \ref{fig:fig_1} with varying DCT magnitude threshold.}
    \label{fig:reconstruct}
\end{figure}

\begin{figure}[htbp]
    \centering
    \includegraphics[width=0.7\linewidth]{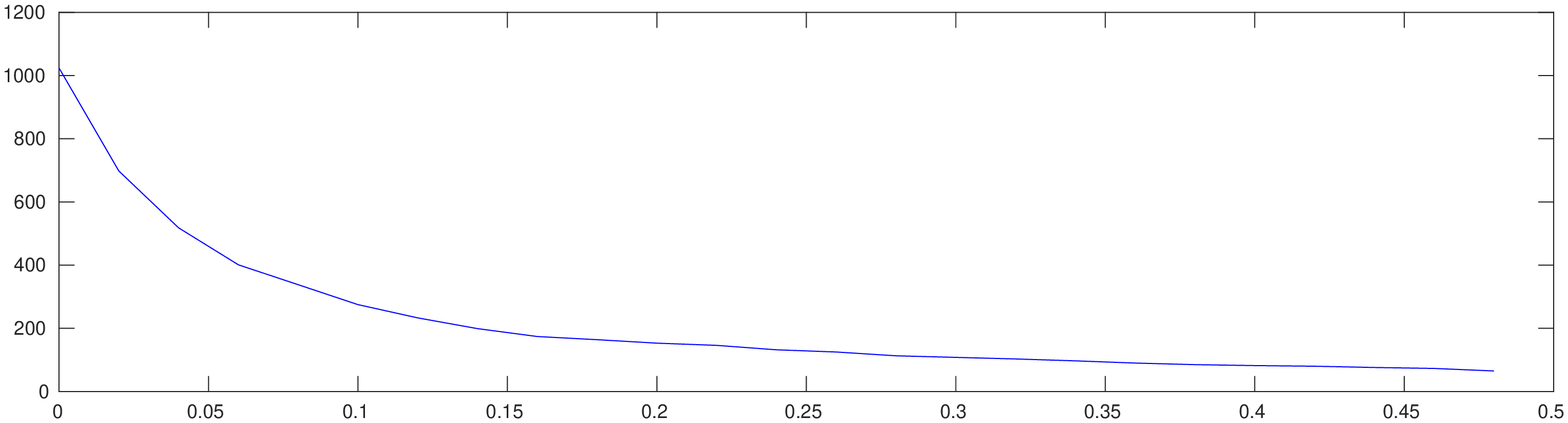} 
    \caption{Number of DCT coefficients v.s. magnitude for the first character in figure \ref{fig:fig_1}.}
    \label{fig:reconN}
\end{figure}

Figure \ref{fig:fig_1} shows a sample form the MNIST data-set together with the DCT of the image patches.
The main attraction of the DCT (as witnessed by its roll in JPEG image compression) is the fact that
for a typical image patch a number of the DCT coefficients are so small they can be ignored,
reducing the dimensionality of the data. Figure \ref{fig:mag_plot} shows a graph of a diagonally rasetered
representation of the 2D DCT plot for the first character in figure \ref{fig:fig_1}.
Figure \ref{fig:reconstruct} shows the effect on the reconstruction of varying a magnitude threshold
filter on the coefficients used to reconstruct the original image patch, the number of remaining
coefficients is graphed in figure \ref{fig:reconN}.
Dropping $30\%$ of the coefficients has almost no noticeable effect on the appearance of the reconstructed character.

Figure \ref{fig:reconstruct} shows the cumulative effect on the reconstruction of a character through increasing
the minimum DCT coefficient magnitude not set to zero. 

\begin{figure}[htbp]
    \centering
    \includegraphics[width=0.7\linewidth]{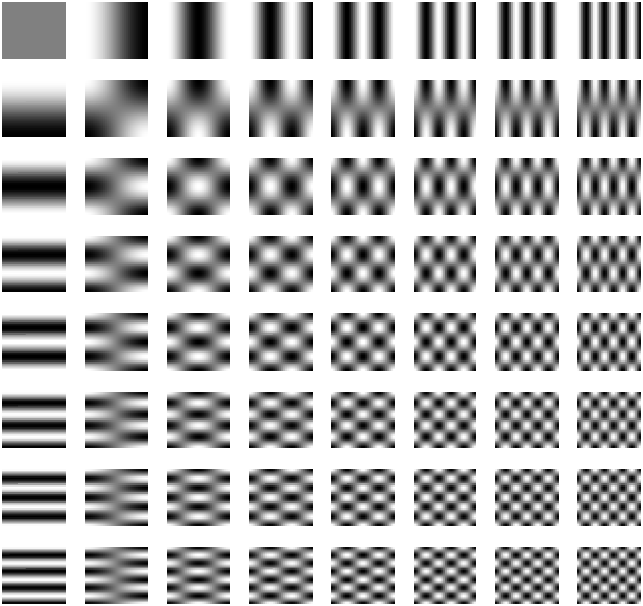} 
    \caption{First few basis vectors of the 32x32 DCT.}
    \label{fig:basis}
\end{figure}

\label{subsec:train_DCT}
A simple feed forward MPL network was used to train a classifier based on the output of the DCT, the network definition is:

\begin{verbatim}
class Net(nn.Module):
    def __init__(self):
        super(Net, self).__init__()
        self.fc1 = nn.Linear(32*32, 350)  
        self.fc2 = nn.Linear(350, 104)
        self.fc3 = nn.Linear(104, 10)

    def forward(self, x):
        x = torch.flatten(x, 1) # flatten all dimensions except the batch dimension
        x = F.relu(self.fc1(x))
        x = F.relu(self.fc2(x))
        x = self.fc3(x)
        return x

criterion = nn.MSELoss()
optimizer = optim.SGD(net.parameters(), lr=0.01)

\end{verbatim}

The network takes in a 32x32 array of DCT coefficients estimated form a 32x32 image patch
(in this case normalised to floating point numbers between 0 and 1).

The performance of the network seemed to be improved by choosing a non zero threshold (in our case 0.02)
on the minimum magnitude of DCT coefficients. The best test performance we achieved was $98.68\%$ 
and to achieve this we had to do a small amount of data augmentation by introducing small
translations (1 pixel shifts) to the training data. This is slightly behind the performance
of leNet at $98.94\%$.

This seems to bear out the notion that CNNs have a modest built in degree of translational tolerance.

\section{Conclusions}
\label{sec:conclusions}

DCT patches offer an attractive way of summarising local appearance. They have been reliably used in
JPEG compression for a generation and seem to provide a more compact summary of image shape than
raw pixel data. The shape information within the DCT coefficients is readily available to a trainable
classifier and achievable performance on MNIST is close to leNet. The basis vectors associated with
DCT coefficients feel similar in certain aspects to the published learned basis vectors of the Visual Transformer.

As a curious aside it is perhaps relevant to note that the human visual system is very good at
spotting right angles. This facility might be related to the use of particular basis
functions in the human visual system.

Using a DCT to estimate linear coefficients of a set of basis functions offers a significant computational
saving over individually learned basis vectors.

There is a small performance difference in favour of the CNN however this is likely because the CNN
learns about edges which are important for human perception i.e. to the system that labeled the training data in the first place.


\bibliographystyle{splncs}
\bibliography{egbib}

\appendix

\section{PyTorch Network Definition}
\label{sec:pytorch_def}

PyTorch \cite{PyTorch_url} was used as the NN training environment. A series of networks
were evaluated with the one used to generate the results in this paper having the
following definition within PyTorch.

\begin{verbatim}
-----dctx32.py----
import torch
import torch.nn as nn
import torch.nn.functional as F

import matplotlib.pyplot as plt
import numpy as np
from PIL import Image

import pdb

class Net(nn.Module):
    def __init__(self):
        super(Net, self).__init__()
        self.fc1 = nn.Linear(32*32, 350)  
        self.fc2 = nn.Linear(350, 104)
        self.fc3 = nn.Linear(104, 10)

    def forward(self, x):
        x = torch.flatten(x, 1) # flatten all dimensions except the batch dimension
        #x = self.dropout(x)
        x = F.relu(self.fc1(x))
        x = F.relu(self.fc2(x))
        x = self.fc3(x)
        return x

    def num_flat_features(self, x):
        size = x.size()[1:]  # all dimensions except the batch dimension
        num_features = 1
        for s in size:
            num_features *= s
        return num_features

net = Net()
criterion = nn.MSELoss()
loss = criterion(output, target)

import torch.optim as optim

# create your optimizer
optimizer = optim.SGD(net.parameters(), lr=0.01)

# in your training loop:
optimizer.zero_grad()   # zero the gradient buffers

print('loading matrices from dusk')
#trn = np.load('trn_MNIST32.npy')
#tst = np.load('tst_MNIST32.npy')

trn = np.load('trn_MNIST32_DCT_02.npy')
tst = np.load('tst_MNIST32_DCT_02.npy')

lbl = np.load('lbl_MNIST32.npy')
lblv = np.load('lblv_MNIST32.npy')
trainL = np.load('trainL_MNIST32.npy')
testL = np.load('testL_MNIST32.npy')

trn = torch.tensor(trn)
tst = torch.tensor(tst)
lbl = torch.tensor(lbl)
lblv = torch.tensor(lblv)
trainL = torch.tensor(trainL)
testL = torch.tensor(testL)

for k in range(80):  
    running_loss = 0.0
    
    for i in range(3200):
        for j in range(16):
            id = np.random.randint(0,numd-1)
            
            # get the inputs
            inputs[j,0] = trn[id]
            labels[j] = lbl[id]

        # zero the parameter gradients
        optimizer.zero_grad()
    
        # forward + backward + optimize
        outputs = net(inputs)
        loss = criterion(outputs, labels)
        loss.backward()
        optimizer.step()

        # print statistics
        running_loss += loss.item()
        if i % 400 == 399:    # print every 200 mini-batches
            print('[%d, %5d] loss: %.3f' %
                  (k + 1, i + 1, running_loss / 20))
            running_loss = 0.0

#now run through EVERY training sample and bundle and resubmit errors.

j = 0
for z in range(0,4):
    for k in range(0, numd-1, 1):  

        input[0,0] = trn[k]
        
        output = net(input)
        x = torch.argmax(output)
        if trainL[k] !=  x.numpy() :       
            inputs[j,0] = trn[k]
            labels[j] = lbl[k]

            j = j + 1
            if j == 16 :
  
                # zero the parameter gradients
                optimizer.zero_grad()
    
                # forward + backward + optimize
                outputs = net(inputs)
                loss = criterion(outputs, labels)
                loss.backward()
                optimizer.step()
                j = 0

print("bundled fails multiplely resubmitted oh great one")

optimizer = optim.SGD(net.parameters(), lr=0.001, momentum=0.9)

shap = trn.shape
nr = shap[2]
nc = shap[3]

#for k in range(30):  
for k in range(1260):  

    running_loss = 0.0
    w = 0;

    for i in range(3200):
        for j in range(16):
            id = np.random.randint(0,numd-1)
            
            # get the inputs
            m = trn[id,0,1:nr-1, 1:nc-1]
            inputs[j,0,:,:] = 0
            
            if w > 11 :
                w = 0

            if w==0:
                inputs[j,0, :nr-2, :nc-2] = m
            if w==1:
                #pdb.set_trace()
                inputs[j,0, :nr-2, 1:nc-1] = m 
            if w==2:
                inputs[j,0, :nr-2, 2:] = m
            if w==3:
                inputs[j,0, 1:nr-1, :nc-2] = m
            if w==4:
                inputs[j,0, 1:nr-1, 2:] = m
            if w==5:
                inputs[j,0, 2:, :nc-2] = m
            if w==6:
                inputs[j,0, 2:, 1:nc-1] = m 
            if w==7:
                inputs[j,0, 2:, 2:] = m 
            if w > 7:
                inputs[j,0] = trn[id]
                
            w = w+1

            #inputs[j,0] = trn[id]
            labels[j] = lbl[id]

        # zero the parameter gradients
        optimizer.zero_grad()
    
        # forward + backward + optimize
        outputs = net(inputs)
        loss = criterion(outputs, labels)
        loss.backward()
        optimizer.step()

        # print statistics
        running_loss += loss.item()
        if i % 400 == 399:    # print every 200 mini-batches
            print('[%d, %5d] loss: %.3f' %
                  (k + 1, i + 1, running_loss / 20))
            running_loss = 0.0


            
print('Finished Training')

#save network weights.
torch.save(net.state_dict(), 'dctx32_MNIST32_dct_002.pth')

#try examples from test set.

numt = tst.shape[0]

input =  torch.tensor(np.zeros(shape=(1,1,32,32))).float()

#pdb.set_trace()


for t in range(1000):
    input[0,0] = tst[t]

    output = net(input)
    x = torch.argmax(output)
    Y = testL[t].cpu().detach().numpy()
    if Y != x.numpy():
        print( t, ':- ', testL[t],  x.numpy())


cnt = 0;
for t in range(numt):
    input[0,0] = tst[t]

    output = net(input)
    x = torch.argmax(output)
    Y = testL[t].cpu().detach().numpy()
    if Y ==  x.numpy() :
        cnt = cnt + 1

print('accuracy: ', cnt*100.0/numt)


\end{verbatim}

The following code was used to make the DCT of the rescaled MNIST data-set.

\begin{verbatim}
--------make_DCT_MNIST_data.py--------
import torch
import idx2numpy
import numpy as np
from torchvision.transforms import ToTensor, Lambda
import matplotlib.pyplot as plt
from PIL import Image
import pdb

import DCTMTX_def

dctmtx = DCTMTX_def.get_dctm() # def. from Mutlab!.

def mnist_tensors():
    
    trainD = idx2numpy.convert_from_file('../MNIST/train-images.idx3-ubyte')
    trainL = idx2numpy.convert_from_file( '../MNIST/train-labels.idx1-ubyte')
    testD = idx2numpy.convert_from_file('../MNIST/t10k-images.idx3-ubyte')
    testL = idx2numpy.convert_from_file('../MNIST/t10k-labels.idx1-ubyte')
    #print(trainD[4])
    print( 'training samples ', trainD.shape[0])
    
    numd = trainD.shape[0]
    s32 = (32,32)
    scal = 1.0/255;

    trainLv = np.zeros(shape=(numd,10))
    trainD32 = np.zeros(shape=(numd,32,32))

        
    #pdb.set_trace()
    
    for d in range(numd):
        img = Image.fromarray(trainD[d]);
        #img = img.resize(s32, Image.ANTIALIAS)
        img = img.resize(s32, Image.LANCZOS)
        m = np.array(img).astype(float)*scal
        dctm = dctmtx @ m @ dctmtx.T
        idx = np.where(abs(dctm) < 0.02 ) 
        dctm[idx] = 0
        trainD32[d] = dctm
        trainLv[d,trainL[d]] = 1
        
    lbl = torch.tensor(trainLv).float()

    #trn = torch.tensor(trainD32*scal).float()
    trn = torch.tensor(trainD32).float()
    
    numv = testD.shape[0]
    testLv = np.zeros(shape=(numv,10))
    testD32 = np.zeros(shape=(numv,32,32))
    
    for v in range(numv):
        img = Image.fromarray(testD[v]);
        #img = img.resize(s32, Image.ANTIALIAS)
        img = img.resize(s32, Image.LANCZOS)
        m = np.array(img).astype(float)*scal
        dctm = dctmtx @ m @ dctmtx.T
        idx = np.where(abs(dctm) < 0.02 ) 
        dctm[idx] = 0

        testD32[v] = dctm
        testLv[v,testL[v]] = 1

        
    lblv = torch.tensor(testLv).float()

    #tst = torch.tensor(testD*scal).float() 
    tst = torch.tensor(testD32).float() 

    
    trn = trn[:,None,:,:]
    tst = tst[:,None,:,:]

    lbl = trainLv[:,None,:]
    lblv = testLv[:,None,:]
    

    
    print(f"Shape of tensor: {trn.shape}")
    print(f"Datatype of tensor: {trn.dtype}")
    print(f"Device tensor is stored on: {tst.device}")
    
    
    #return( trn, tst, lbl, lblv, trainL, testL )
    return( trn, tst, trainLv, testLv, trainL, testL )



(trn, tst, lbl, lblv, trainL, testL ) = mnist_tensors()
#pdb.set_trace()
df = 9

np.save('trn_MNIST32_DCT_03.npy', trn)
np.save('tst_MNIST32_DCT_03.npy', tst)

if 0 :
    np.save('lbl_MNIST32.npy', lbl)
    np.save('lblv_MNIST32.npy', lblv)
    np.save('trainL_MNIST32.npy', trainL)
    np.save('testL_MNIST32.npy', testL)


\end{verbatim}

\end{document}